\documentclass[11pt]{article}

\usepackage{aaai}
\usepackage{url}

\usepackage{times}
\usepackage{latexsym}
\usepackage{tcolorbox}
\usepackage{makecell}
\usepackage{array}
\usepackage{booktabs}
\usepackage{multirow}
\usepackage{makecell}
\usepackage{listings}
\usepackage{longtable}
\usepackage{newunicodechar}
\newunicodechar{≥}{\ensuremath{\geq}}
\usepackage{amsmath}

\usepackage{booktabs, makecell, multirow, graphicx}

\usepackage[T1]{fontenc}

\usepackage[utf8]{inputenc}

\usepackage{microtype}

\usepackage{inconsolata}

\usepackage{graphicx}

\title{BAID: A Benchmark for Bias Assessment of AI Detectors}

\author{Priyam Basu \\
  Superhuman \\
  \texttt{priyam.basu@grammarly.com} \\\And
  Yunfeng Zhang \\
  Superhuman \\
  \texttt{yunfeng.zhang@grammarly.com} \\\And
  Vipul Raheja \\
  Superhuman \\
  \texttt{vipul.raheja@grammarly.com} \\}

\begin{document}
\maketitle
\begin{abstract}
AI-generated text detectors have recently gained adoption in educational and professional contexts. Prior research has uncovered isolated cases of bias, particularly against English Language Learners (ELLs) however, there is a lack of systematic evaluation of such systems across broader sociolinguistic factors. In this work, we propose BAID, a comprehensive evaluation framework for AI detectors across various types of biases. As a part of the framework, we introduce over 200k samples spanning 7 major categories: demographics, age, educational grade level, dialect, formality, political leaning, and topic. We also generated synthetic versions of each sample with carefully crafted prompts to preserve the original content while reflecting subgroup-specific writing styles. Using this, we evaluate four open-source state-of-the-art AI text detectors and find consistent disparities in detection performance, particularly low recall rates for texts from underrepresented groups. Our contributions provide a scalable, transparent approach for auditing AI detectors and emphasize the need for bias-aware evaluation before these tools are deployed for public use.
\end{abstract}

\section{Introduction}
As large language models (LLMs) such as GPT-4 \cite{openai2024gpt4technicalreport} and LLaMA \cite{touvron2023llamaopenefficientfoundation} continue to improve, the line between machine-generated and human-written text is becoming increasingly difficult to draw. These models now produce writing that is not only grammatically correct but also stylistically sophisticated and contextually nuanced \cite{brown2020languagemodelsfewshotlearners}, while being indistinguishable to the amateur eye. Recent advancements have introduced new risks around the generation of deceptive content, raising serious concerns about their potential to mislead or manipulate public perception \cite{solaiman2019releasestrategiessocialimpacts}. These risks span a range of real-world applications, including the automated creation of fabricated news stories \cite{zellers2020defendingneuralfakenews}, fake product reviews \cite{meng2025largelanguagemodelshidden}, inauthentic social media posts intended to influence public opinion \cite{loth2024blessingcursesurveyimpact} as well as phishing attacks \cite{thapa2025phishingdetectiongenaiera}. In parallel, educators have expressed growing unease over the use of generative tools in academic settings \cite{CURRIE2023719}.

Recent works have proposed a variety of detection methods aimed at distinguishing machine-written text from human-written text. These efforts span a range of approaches, from leveraging statistical irregularities in generated outputs \cite{gehrmann2019gltrstatisticaldetectionvisualization} to training supervised classifiers on curated datasets \cite{mitchell2023detectgptzeroshotmachinegeneratedtext}. Most detectors operate under a binary assumption that a given input is either fully AI-generated or fully human-written. This implies they evaluate the input text at a paragraph or document level, while some work focusses on fine-grained detection, including phrase-level or even token-level classification \cite{teja2025finegraineddetectionaigeneratedtext}.

Although significant progress has been made in developing and evaluating AI-generated text detectors, these models have not been tested for fairness and equity. In particular, research on bias in AI detectors remains sparse. \cite{liang2023gptdetectorsbiasednonnative} systematically investigated this issue, where they found that widely-used detectors disproportionately classify texts written by non-native English speakers as AI-generated due to their lower linguistic perplexity. This discovery underscores a troubling consequence: detectors may inadvertently penalize individuals based on their language background, even when their writing is entirely original. Motivated by this insight, our work extends the investigation of bias in AI detectors by evaluating their behavior across a broader and more diverse set of dimensions. Specifically, we examine seven types of bias - demographics, age, educational grade level, dialect, formality, political leaning and topic, to offer a more comprehensive assessment of how detection systems may fail across different groups. By doing so, we aim to highlight not only the technical limitations of current detectors but also the social implications of deploying them at scale without rigorous fairness evaluations.

\section{Related Works}
Various methods have been developed for the detection of AI-generated text. Early approaches like \cite{gehrmann2019gltrstatisticaldetectionvisualization} used statistical cues and visualizations to exploit the fact that AI-generated text often relies on a narrower range of high-probability word patterns. Other methods like \cite{bao2024fastdetectgptefficientzeroshotdetection} provide zero-shot way solutions by analyzing outputs via perplexity or entropy differences. ZipPy \cite{thinkst2023ZipPy} foregoes heavy neural networks for speed by using compression ratios to measure textual novelty (an indirect perplexity metric) against a reference corpus of AI-generated text. However, more recent works like GPTZero \cite{mitchell2023detectgptzeroshotmachinegeneratedtext}, Desklib \cite{desklib2025aitextdetector} focus on this as a finetuning task on human vs ai-generated texts. Beyond purely AI- or human-authored texts, researchers have started examining hybrid human-AI texts. \cite{zeng2024detectingaigeneratedsentenceshumanai} explore sentence-level detection in collaborative writing, highlighting that identifying AI-generated segments amid human revisions is extremely challenging. They found that when humans selectively edit or intermix AI-generated sentences, detectors struggle due to rapidly switching authorship and minimal stylistic cues in short segments. This has lead to some more works on sentence-level and phrase-level AI detection \cite{wang2023seqxgptsentencelevelaigeneratedtext}.

Recent benchmarking efforts have focused on systematically evaluating the quality and generalization of AI text detectors across domains, models, and use cases. \cite{pudasainibenchmarking} highlight that detectors often fail under distribution shifts, paraphrasing, and newer model generations, highlighting the brittleness of current approaches. \cite{yu2025paperreviewedllmbenchmarking} examine the effectiveness of detectors in academic review scenarios and reveal substantial false-positive risks when evaluating legitimate human writing, especially in specialized or formal domains. \cite{tao2024reliabledetectionllmgeneratedtexts} demonstrated that detection performance varies widely across languages and content genres. \cite{dugan2024raidsharedbenchmarkrobust}, showed significant degradation under paraphrased or obfuscated text conditions.

With more widespread use of these detectors, concerns have risen around their reliability and bias. \cite{liang2023gptdetectorsbiasednonnative} in their paper ran an experiment which showed how GPT-detectors were extremely biased against non-native English speakers, incorrectly classifying more than half of the TOEFL human-written essays by English Language Learner (ELL) students as LLM generated. This stems from the underlying low perplexity values (inverse of word sequence probability) of essays written by ELL students. \cite{chu2024nationalityraceethnicitybiases}] discuss how source heuristics such as nationality and content heuristics like linguistics attributes play a significant factor in authenticity detection. They further talk about about content from Asian and Hispanic writers are more likely to be judged as AI users when
labeled as domestic students, suggesting interactions between racial stereotypes and AI detection, even when judged by humans. However, they do not run any experiments or show any empirical results to prove this hypothesis, a gap which we explore in this work.

\section{Dataset}

\subsection{Overview}
We introduce BAID, a benchmark designed to evaluate the fairness of AI-generated text detectors across diverse demographic and linguistic subgroups. While existing benchmarks primarily assess detector performance under standard or neutral conditions, our benchmark evaluates fairness in conditions where subgroup attributes may influence detector outputs.

To build the dataset, we collect multiple human-written documents spanning seven major bias groups, each containing multiple subgroups. For every human-written document, we generate a corresponding AI-written version using LLMs prompted with carefully crafted instructions that control for a human-like tone. This ensures that both human and AI texts share comparable semantic contexts while differing in authorship. The resulting dataset consists of three fields: \texttt{human\_written\_document}, \texttt{AI\_generated\_document}, and \texttt{subgroup\_value}.

In contrast to existing datasets that evaluate models under typical or neutral inputs, BAID emphasizes fairness evaluation under bias-revealing conditions. Similar to prior works such as FLEX \cite{jung2025flexbenchmarkevaluatingrobustness}, which test language models in extreme fairness scenarios, BAID seeks to expose potential disparities in how AI detectors behave across population subgroups.

The design of BAID follows three guiding principles - (a) Fairness coverage: The dataset should include a wide range of demographic and linguistic variables that reflect real-world diversity. (b) Semantic control: Human and AI texts should express the same content and intent, ensuring that fairness differences arise from subgroup attributes rather than topical or stylistic drift. (c) Practical evaluation: BAID is intended as a diagnostic tool to evaluate model fairness across a variety of writing domains, from formal essays to conversational text.

\subsection{Bias Dimensions}
While earlier work (e.g., Stanford HAI \cite{liang2023gptdetectorsbiasednonnative}) focused on the disadvantage faced by English Language Learner (ELL) students, BAID broadens this scope to include demographic, social, and stylistic dimensions. Each bias type represents a fairness-relevant variable where detector disparities could translate into real-world harms.

\begin{itemize}
    \item \textbf{Demographic bias.} We use the ASAP 2.0 dataset \cite{scrosseye2020asap2}, which contains persuasive essays from standardized writing assessments. Metadata includes author race/ethnicity, gender, socioeconomic status, disability status, and ELL status.
    \item \textbf{Age bias.} Based on the Blog Authorship Corpus \cite{rtatman2020blogauthorship}, which includes 600K posts from roughly 19,000 bloggers aged 13–48, grouped into four ranges: teens, 20s, 30s, and 40s.
    \item \textbf{Grade-level bias.} The ASAP 2.0 corpus also includes grade-level information ranging from 8 to 12, allowing comparison of writing maturity effects on detector outcomes.
    \item \textbf{Dialect bias.} We examine three English varieties which include: African American Vernacular English (AAVE) \cite{blodgett2016demographic}, Singaporean English (Singlish) \cite{rtatman2017singlish}, and Standard American English (SAE) \cite{groenwold-etal-2020-investigating} to measure robustness across dialectal variation.
    \item \textbf{Formality bias.} We use the GenZ vs. Standard English dataset \cite{seraaphonano2024formal_informal}, which contains 820 paired sentences contrasting informal GenZ phrasing with formal equivalents.
    \item \textbf{Topic bias.} Using the Blog Authorship Corpus, we select ten major topical categories (Arts, Communication/Media, Education, Engineering, Internet, Law, Non-profit, Student, Technology, and Unknown) to test whether detector fairness depends on subject matter.
    \item \textbf{Political ideology bias.} We adopt the dataset from \cite{baly2020we}, which includes articles annotated as left-leaning, neutral, or right-leaning, to evaluate ideological sensitivity.
\end{itemize}

\begin{table*}[t]
\centering
\scriptsize
\renewcommand\arraystretch{0.82}
\setlength{\tabcolsep}{8pt}

\begin{tabular}{m{1.5cm}m{3.5cm}p{3.5cm}m{1.0cm}}
\toprule
\textbf{Bias Type} & \textbf{Subgroup} & \textbf{Dataset Citation} & \textbf{Count} \\
\midrule

\multirow[t]{19}{*}{Demographic}
  & Race/ethnicity:                                 & \multirow[t]{19}{*}{\cite{scrosseye2020asap2}} & \\
  & \quad American Indian/Alaskan Native            &  & 184  \\
  & \quad Asian/Pacific Islander                    &  & 2988 \\
  & \quad Black/African American                    &  & 3800 \\
  & \quad Hispanic/Latino                           &  & 3192 \\
  & \quad Two or more races/Other                   &  & 2048 \\
  & \quad White                                     &  & 4000 \\
  & ELL status:                                     &  &  \\
  & \quad Yes                                       &  & 3996 \\
  & \quad No                                        &  & 4000 \\
  & Student disability status:                      &  &  \\
  & \quad Yes                                       &  & 3996 \\
  & \quad No                                        &  & 3400 \\
  & Socioeconomic status:                           &  &  \\
  & \quad Economically disadvantaged                &  & 3880 \\
  & \quad Not economically disadvantaged            &  & 3996 \\
  & Gender:                                         &  &  \\
  & \quad Female                                    &  & 4000 \\
  & \quad Male                                      &  & 3996 \\
\midrule

\multirow[t]{5}{*}{Grade level}
  & Grade 8   & \multirow[t]{5}{*}{\cite{scrosseye2020asap2}} & 1950 \\
  & Grade 9   &  & 52   \\
  & Grade 10  &  & 426  \\
  & Grade 11  &  & 2754 \\
  & Grade 12  &  & 2638 \\
\midrule

\multirow[t]{4}{*}{Age level}
  & Teens  & \multirow[t]{4}{*}{\cite{rtatman2020blogauthorship}} & 4991 \\
  & 20s    &  & 4996 \\
  & 30s    &  & 4996 \\
  & 40s    &  & 4985 \\
\midrule

\multirow[t]{3}{*}{Dialect}
  & AAVE      & \multirow[t]{3}{*}{\makecell[c]{\cite{blodgett2016demographic}\\\cite{rtatman2017singlish}\\\cite{groenwold-etal-2020-investigating}}} & 19180 \\
  & Singlish  &  & 10000 \\
  & SAE       &  & 8070  \\
\midrule

\multirow[t]{2}{*}{Formality bias}
  & GenZ English      & \multirow[t]{2}{*}{\cite{seraaphonano2024formal_informal}} & 3280 \\
  & Standard English  &  & 3280 \\
\midrule

\multirow[t]{10}{*}{Topic level}
  & Arts                 & \multirow[t]{10}{*}{\cite{rtatman2020blogauthorship}} & 4995 \\
  & Communication/Media  &  & 4998 \\
  & Education            &  & 4995 \\
  & Engineering          &  & 4990 \\
  & Internet             &  & 4984 \\
  & Law                  &  & 7487 \\
  & Non-profit           &  & 4995 \\
  & Student              &  & 4994 \\
  & Technology           &  & 4998 \\
  & Unknown              &  & 4154 \\
\midrule

\multirow[t]{3}{*}{Political leaning}
  & Left     & \multirow[t]{3}{*}{\cite{baly2020we}} & 12800 \\
  & Neutral  &  & 11010 \\
  & Right    &  & 13722 \\
\bottomrule
\end{tabular}
\caption{Dataset composition across bias dimensions, subgroups, and sample counts}
\label{tab:bias-datasets}
\end{table*}

\subsection{Prompt Design}
To generate "AI-authored" counterparts, we use a set of structured zero-shot prompts that simulate light-touch human revisions. For essays and articles, the model is prompted to act as an editor, rewriting overly formal or robotic phrases while preserving paragraph structure and meaning. Prompts explicitly discourage stereotypical AI markers such as "in this essay," "delve into," or "in conclusion," and instead promote natural discourse connectors like "so," "but," and "also."  

For short-form or conversational inputs (e.g., tweets and messages), prompts are customized to match the linguistic features of each dialect. For example, AAVE samples emphasize authentic syntactic and lexical constructions, while Singlish prompts incorporate pragmatic particles and colloquial phrasing. In all cases, the models are instructed to maintain semantic similarity to the human-written text. The prompt template is provided in the Appendix

\subsection{Generation Process}
All AI-generated documents are produced using GPT-4.1 \cite{openai2024gpt4technicalreport} and Claude Sonnet 3.7 \cite{anthropic2025claude3dot7}. We use a multi-threaded generation pipeline with built-in retry mechanisms to handle rate-limit and timeout errors. Generated outputs are filtered for completeness and cleaned to remove hashtags, emojis and links. Each generation is paired with its corresponding human-written sample and subgroup label.

\subsection{Data Quality and Filtering}
To ensure quality and reliability, we apply a multi-stage validation process:
\begin{itemize}
    \item \textbf{Automatic filtering:} We discard samples with token repetition and incomplete generations.
    \item \textbf{Semantic alignment:} Using sentence-level embeddings, we compute cosine similarity between human and AI pairs to confirm that the generated text preserves core meaning with a threshold of 0.85.
\end{itemize}

This process ensures that fairness measurements reflect true subgroup differences rather than artifacts of poor generation quality. The final dataset contains 208166 document pairs distributed across seven bias types and 41 subgroups. Table~\ref{tab:bias-datasets} illustrates the different types of biases, subgroups and count.

\section{Detectors}
We evaluate two types of detectors: neural models and statistical models. In total, we apply four widely used AI-generated text detectors on the BAID benchmark:
\begin{itemize}
    \item \textbf{Desklib} \cite{desklib2025aitextdetector} - A model developed by fine-tuning a deberta-v3-large \cite{he2023debertav3improvingdebertausing} model with adversarial attacks across different domains.
    \item \textbf{E5-small} \cite{mayzhou2024e5small} - A lightweight model built using LoRA \cite{hu2021loralowrankadaptationlarge} fine-tuning on the E5-small \cite{wang2024multilinguale5textembeddings} encoder model.
    \item \textbf{Radar} \cite{hu2023radarrobustaitextdetection} - A model jointly trained on a detector and a paraphraser task via adversarial learning to improve resilience against LLM-based paraphrasing and cross-model transfer.
    \item \textbf{ZipPy} \cite{thinkst2023ZipPy} - A fast detection model that correlates ease of compression to perplexity as a metric for authorship authenticity.
\end{itemize}

We run each detector over the entire dataset, individually for human-written and ai-generated text, and compare performance across different biases and subgroups. We use the default decision thresholds for each detector, treating them as black-box systems.

\section{Evaluation}
Table~\ref{tab:all-bias} reports per-subgroup results for each bias. For a fair evaluation, we evaluate the detectors only on the human-written texts, as any observed bias would meaningfully originate from human authorship. In contrast, AI-generated texts merely simulate subgroup characteristics through prompting and do not truly represent the underlying demographic or linguistic identity, making them unsuitable for assessing fairness across real-world subgroups. The metrics we focus on are precision, recall and F1. Three findings are consistent:

\begin{table*}[t]
\centering
\scriptsize 
\renewcommand{\arraystretch}{1.5}
\renewcommand\arraystretch{0.70}
\setlength{\tabcolsep}{5pt}

\begin{tabular}{l l *{3}{c} *{3}{c} *{3}{c} *{3}{c}}
\toprule
\multirow{2}{*}{\makecell[l]{Dimension}} &
\multirow{2}{*}{\makecell[l]{Subgroup}} &
\multicolumn{3}{c}{\textbf{Desklib}} &
\multicolumn{3}{c}{\textbf{E5}} &
\multicolumn{3}{c}{\textbf{Radar}} &
\multicolumn{3}{c}{\textbf{ZipPy}} \\
\cmidrule(lr){3-5}\cmidrule(lr){6-8}\cmidrule(lr){9-11}\cmidrule(lr){12-14}
 & & P & R & F1 & P & R & F1 & P & R & F1 & P & R & F1 \\
\midrule

\multirow{2}{*}{Gender}
  & F & 0.98 & 0.84 & 0.91 & 0.98 & 0.17 & 0.29 & 0.60 & 0.66 & 0.62 & 0.24 & 0.15 & 0.20 \\
  & M & 0.99 & 0.85 & 0.92 & 0.99 & 0.25 & 0.40 & 0.61 & 0.64 & 0.62 & 0.23 & 0.16 & 0.19 \\
\cmidrule(lr){1-14}

\multirow{6}{*}{Race}
  & American Indian  & 0.97 & 0.65 & 0.78 & 0.89 & 0.17 & 0.30 & 0.54 & 0.61 & 0.57 & 0.19 & 0.13 & 0.15 \\
  & API  & 0.98 & 0.82 & 0.90 & 0.99 & 0.19 & 0.31 & 0.55 & 0.64 & 0.59 & 0.24 & 0.15 & 0.19 \\
  & African American & 0.98 & 0.87 & 0.93 & 0.98 & 0.29 & 0.45 & 0.63 & 0.66 & 0.64 & 0.28 & 0.21 & 0.24 \\
  & Hispanic         & 0.99 & 0.85 & 0.91 & 0.99 & 0.23 & 0.37 & 0.61 & 0.65 & 0.63 & 0.25 & 0.18 & 0.21 \\
  & White            & 0.98 & 0.86 & 0.92 & 0.97 & 0.21 & 0.34 & 0.60 & 0.65 & 0.62 & 0.24 & 0.17 & 0.20 \\
  & Two or more      & 0.99 & 0.84 & 0.91 & 0.95 & 0.21 & 0.34 & 0.60 & 0.70 & 0.65 & 0.26 & 0.21 & 0.23 \\
\cmidrule(lr){1-14}

\multirow{2}{*}{Economic Disadvantage}
  & Y & 0.98 & 0.84 & 0.90 & 0.97 & 0.27 & 0.43 & 0.60 & 0.65 & 0.62 & 0.26 & 0.20 & 0.23 \\
  & N & 0.98 & 0.83 & 0.90 & 0.98 & 0.20 & 0.33 & 0.59 & 0.65 & 0.61 & 0.22 & 0.16 & 0.19 \\
\cmidrule(lr){1-14}

\multirow{2}{*}{Disability Status}
  & Y & 0.98 & 0.81 & 0.89 & 0.98 & 0.19 & 0.32 & 0.62 & 0.65 & 0.63 & 0.29 & 0.26 & 0.27 \\
  & N & 0.99 & 0.84 & 0.91 & 0.98 & 0.37 & 0.54 & 0.60 & 0.66 & 0.63 & 0.23 & 0.15 & 0.18 \\
\cmidrule(lr){1-14}

\multirow{2}{*}{ELL Status}
  & Y & 0.97 & 0.77 & 0.86 & 0.99 & 0.20 & 0.32 & 0.61 & 0.63 & 0.62 & 0.23 & 0.17 & 0.20 \\
  & N & 0.99 & 0.85 & 0.92 & 0.97 & 0.24 & 0.45 & 0.61 & 0.65 & 0.63 & 0.29 & 0.23 & 0.25 \\
\cmidrule(lr){1-14}

\multirow{5}{*}{Grade Level}
  & 8  & 0.98 & 0.89 & 0.93 & 0.99 & 0.23 & 0.38 & 0.72 & 0.57 & 0.63 & 0.10 & 0.34 & 0.05 \\
  & 9  & 0.99 & 0.99 & 0.99 & 0.99 & 0.27 & 0.43 & 0.76 & 0.73 & 0.75 & 0.58 & 0.55 & 0.56 \\
  & 10 & 0.99 & 0.90 & 0.94 & 0.99 & 0.23 & 0.37 & 0.74 & 0.75 & 0.74 & 0.06 & 0.02 & 0.04 \\
  & 11 & 0.99 & 0.92 & 0.95 & 0.99 & 0.18 & 0.30 & 0.72 & 0.67 & 0.70 & 0.08 & 0.03 & 0.04 \\
  & 12 & 0.99 & 0.93 & 0.96 & 0.99 & 0.26 & 0.41 & 0.75 & 0.73 & 0.74 & 0.08 & 0.02 & 0.03 \\
\cmidrule(lr){1-14}

\multirow{4}{*}{Age Level}
  & Teens & 0.64 & 0.92 & 0.76 & 0.58 & 0.55 & 0.57 & 0.27 & 0.31 & 0.29 & 0.49 & 0.95 & 0.65 \\
  & 20s   & 0.65 & 0.88 & 0.75 & 0.63 & 0.39 & 0.48 & 0.23 & 0.22 & 0.23 & 0.50 & 0.97 & 0.66 \\
  & 30s   & 0.68 & 0.86 & 0.76 & 0.65 & 0.32 & 0.43 & 0.28 & 0.26 & 0.27 & 0.50 & 0.96 & 0.66 \\
  & 40s   & 0.68 & 0.80 & 0.74 & 0.66 & 0.28 & 0.39 & 0.30 & 0.26 & 0.28 & 0.50 & 0.96 & 0.66 \\
\cmidrule(lr){1-14}

\multirow{3}{*}{Dialect Bias}
& Singlish & 0.44 & 0.26 & 0.33 & 0.28 & 0.35 & 0.31 & 0.18 & 0.25 & 0.21 & 0.49 & 0.98 & 0.66 \\
  & AAVE     & 0.38 & 0.20 & 0.27 & 0.42 & 0.71 & 0.52 & 0.30 & 0.52 & 0.38 & 0.50 & 0.98 & 0.66 \\
  & SAE  & 0.74 & 0.35 & 0.47 & 0.50 & 0.97 & 0.66 & 0.32 & 0.72 & 0.44 & 0.50 & 0.99 & 0.67 \\
\cmidrule(lr){1-14}

\multirow{2}{*}{Formality Bias}
  & GenZ     & 0.16 & 0.12 & 0.14 & 0.04 & 0.04 & 0.04 & 0.01 & 0.04 & 0.02 & 0.50 & 0.99 & 0.67 \\
  & Standard & 0.52 & 0.41 & 0.46 & 0.72 & 0.55 & 0.62 & 0.38 & 0.30 & 0.33 & 0.54 & 0.97 & 0.70 \\
\cmidrule(lr){1-14}

\multirow{10}{*}{Topic Level}
  & Arts                & 0.69 & 0.89 & 0.78 & 0.60 & 0.46 & 0.52 & 0.28 & 0.25 & 0.26 & 0.50 & 0.97 & 0.65 \\
  & Communication/Media & 0.73 & 0.85 & 0.79 & 0.63 & 0.35 & 0.45 & 0.29 & 0.24 & 0.27 & 0.50 & 0.97 & 0.66 \\
  & Education           & 0.73 & 0.88 & 0.80 & 0.66 & 0.39 & 0.49 & 0.32 & 0.28 & 0.30 & 0.50 & 0.95 & 0.66 \\
  & Engineering         & 0.72 & 0.88 & 0.79 & 0.64 & 0.44 & 0.52 & 0.32 & 0.29 & 0.30 & 0.50 & 0.97 & 0.66 \\
  & Internet            & 0.68 & 0.86 & 0.76 & 0.62 & 0.34 & 0.43 & 0.29 & 0.24 & 0.26 & 0.50 & 0.96 & 0.66 \\
  & Law                 & 0.79 & 0.89 & 0.83 & 0.66 & 0.32 & 0.43 & 0.29 & 0.24 & 0.26 & 0.51 & 0.98 & 0.67 \\
  & Non-profit          & 0.75 & 0.88 & 0.81 & 0.72 & 0.44 & 0.55 & 0.31 & 0.22 & 0.25 & 0.50 & 0.97 & 0.66 \\
  & Student             & 0.68 & 0.90 & 0.77 & 0.59 & 0.51 & 0.55 & 0.28 & 0.28 & 0.28 & 0.49 & 0.96 & 0.65 \\
  & Technology          & 0.71 & 0.86 & 0.78 & 0.64 & 0.38 & 0.48 & 0.27 & 0.23 & 0.25 & 0.50 & 0.97 & 0.66 \\
  & Unknown             & 0.70 & 0.87 & 0.78 & 0.66 & 0.41 & 0.51 & 0.30 & 0.26 & 0.28 & 0.49 & 0.96 & 0.65 \\
\cmidrule(lr){1-14}

\multirow{3}{*}{Political Ideology}
  & Left leaning  & 0.98 & 0.93 & 0.96 & 0.73 & 0.06 & 0.11 & 0.51 & 0.99 & 0.68 & 0.45 & 0.81 & 0.58 \\
  & Neutral       & 0.99 & 0.89 & 0.93 & 0.57 & 0.03 & 0.06 & 0.51 & 0.99 & 0.68 & 0.46 & 0.83 & 0.59 \\
  & Right leaning & 0.99 & 0.95 & 0.97 & 0.78 & 0.08 & 0.14 & 0.51 & 0.99 & 0.68 & 0.45 & 0.82 & 0.58 \\
\bottomrule
\end{tabular}
\caption{Performance by subgroup across all bias dimensions on human-written texts}
\label{tab:all-bias}
\end{table*}

\paragraph{Precision.}
Across all bias dimensions, Desklib demonstrated high precision (0.97-0.99) for demographic, grade-level, and political subgroups, confirming its strong reliability in identifying AI-generated text when present. However, its precision declined on dialectal and informal writing (e.g., 0.44 for Singlish and 0.16 for GenZ). E5 also achieves consistently high precision on demographics and grade levels (0.95-0.99), though its performance drops for dialectal text (0.28-0.50) and some topic categories (0.60-0.75), with notably poor precision for GenZ content (0.04). Radar maintains stable, mid-range precision values (0.55-0.76) across subgroups, suggesting moderate but consistent performance. In contrast, ZipPy records the lowest precision on demographic and grade-level data (0.19-0.31) but achieves moderately better precision (0.49-0.54) on dialect, formality, and topic biases. Overall, the results indicate that neural models such as Desklib and E5 handle precision more consistently, whereas compression-based systems like ZipPy exhibit higher sensitivity to writing style and input length.

\paragraph{Recall.}
Desklib achieves strong recall on demographic and grade-level dimensions (0.83-0.96), demonstrating robustness to linguistic diversity, though it declines on dialectal and informal text (0.12-0.35). E5 records very low recall for demographic and political groups (0.03-0.45) but substantially higher recall for dialects such as Singlish and Standard English (0.35-0.97 and 0.55, respectively). Radar remains relatively balanced (0.57-0.72) but shows reduced recall on dialectal and informal categories. ZipPy performs worst overall, with recall collapsing on demographics and grade-level data (0.02-0.55) but reaching extremely high levels (0.95-0.99) for age, dialect, topic, and GenZ subgroups. These results highlight the importance of recall in fairness-sensitive applications: detectors that underperform in recall risk misclassifying human-written text as AI-generated, disproportionately penalizing certain undunderrepresentedoups. Consequently, improving recall robustness across linguistic variation remains an essential goal for fair detection.

\paragraph{F1.}
Desklib performs the most consistently across all bias dimensions, with high F1 scores ranging from 0.89 to 0.96 on demographic and grade-level subgroups. Its performance decreases, however, on dialectal and informal registers (0.14-0.47). Radar follows with moderate F1 scores (0.60-0.75), though it also experiences noticeable drops on dialect and formality dimensions (0.21-0.33). Despite its high precision, E5 yields relatively low F1 values for demographics (0.30-0.45) and GenZ English (0.04), improving only for Standard English (0.66) and certain topical categories (0.44-0.55). ZipPy exhibits the weakest overall performance, with very low F1 on demographics and grade levels (0.03-0.27), but achieves higher values (0.65-0.70) on dialectal, formality, and topical text. These findings underscore that aggregate F1 averages can mask substantial subgroup disparities, reinforcing the need for disaggregated fairness evaluations.

\subsection{Length Sensitivity in ZipPy}
ZipPy relies on compression-based heuristics rather than supervised training. The detector "seeds" a compression stream with AI-generated text and then measures how efficiently new samples compress relative to that seed. TextTexts that share lexical or structural similarity with AI-generated ones achieve higher compression ratios, while human-written texts are typically more variable and compress less efficiently.ever, ZipPy’s effectiveness is highly sensitive to input length, shorter texts offer fewer repeating tokens and thus less reliable compression estimates.

\section{Limitations}
While we provide a benchmark across a variety of domains, our analysis is limited to a fixed set of only four detectors. Including larger commercial systems or emerging hybrid detectors (e.g multimodal or cross-lingual models) could provide additional insight into fairness trends. Another limitation is that even though it spans seven bias dimensions, the focus is on English text, which can be extended to other languages to enable multilingual evaluation. Finally, ZipPy and other statistical detectors are highly sensitive to input length and formatting, while neural detectors may be influenced by pretraining corpora biases. These architectural differences complicate direct comparisons, which call for including multi-language corpora, dynamic threshold calibration, and experiments with hybrid detection models to better understand fairness under broader real-world conditions.

\section{Conclusion}
In this work, we introduced a benchmark designed to systematically evaluate the fairness of AI-generated text detectors across diverse demographic and linguistic subgroups. We revealed consistent disparities in detection behavior, most notably recall gaps that disproportionately penalize under-represented writing styles such as dialectal and informal English. While neural detectors like Desklib maintain high overall accuracy, statistical systems such as ZipPy exhibit length-dependent variability, underscoring how detector architecture and input characteristics jointly shape fairness outcomes. Our findings highlight that existing detectors, though effective in aggregate metrics, exhibit bias. This emphasizes the need for bias-aware auditing, training data diversity, and model calibration in AI detection research.  
We hope BAID will serve as a foundation for developing more equitable and transparent detection systems that perform reliably across different types of population, writing styles, and contexts.

\bibliographystyle{aaai}
\bibliography{references}

\section{Appendix}
\subsection{Evaluation of AI-Generated Texts}
We also ran the same subgroup-level evaluation on the AI-generated samples of the BAID dataset. However, it is important to note that these results do not reflect inherent bias in the same way as the human-written samples. Each AI-generated text was produced using prompts that explicitly instructed the model to rewrite an existing human-written document from the perspective of the original author’s subgroup. As such, any linguistic or stylistic variation in these samples is a result of the generation process rather than genuine demographic or experiential differences. Consequently, the subgroup-level patterns observed in this analysis should be interpreted cautiously, as they reflect prompt conditioning of the language models and quality analysis rather than authentic subgroup bias.

Across most dimensions, detectors show noticeably higher recall on AI-generated text than on human-written samples, suggesting that synthetic outputs still carry the statistical fingerprints of machine generation. Among all systems, Desklib stands out for its consistency since it maintains high precision (0.8-0.9) and recall above 0.97 across subgroups, leading to F1 scores well over 0.9. In other words, Desklib reliably identifies generated text regardless of subgroup conditioning. E5 also achieves very high recall but at the expense of precision (0.55-0.60), indicating a tendency to predict AI falsely, an expected trade-off for detectors tuned toward recall. Radar, which incorporates adversarial training and paraphrase modeling, performs more unevenly, with F1 scores typically in the 0.6-0.7 range and weaker results on stylistic dimensions such as dialect and formality. This pattern suggests that Radar’s adversarial robustness does not fully capture stylistic or prompt-induced variation. Finally, ZipPy, the compression-based detector, behaves less predictably. It achieves very high recall (above 0.9) on longer, more regular texts but suffers from low precision (0.3-0.5) and inconsistent F1 scores across subgroups, reflecting its sensitivity to text length and lexical repetition.

Overall, this demonstrates that fairness metrics computed on AI-generated text primarily capture model calibration and sensitivity to surface-level linguistic properties, not representational bias. The uniformly high recall across subgroups confirms that detectors reliably recognize the statistical regularities of generated text, while small subgroup variations reveal the influence of prompt structure and lexical complexity. These results assess how detectors generalize across controlled synthetic variations, rather than how they behave toward real human diversity.

\subsection{Prompt Template}
\label{tab:Prompt Template}
\begin{tcolorbox}[colback=gray!5,colframe=gray!50!black,title={Prompt Template for GenZ Rewriter}]
\small
\textbf{SYSTEM\_PROMPT\_GENZ} \\[3pt]
You are an AI rewriter that transforms and paraphrases existing text messages into Gen Z style.

Your job is to rewrite a piece of text to sound like it was written in the casual, slang-filled, playful, and internet-savvy style of Gen Z online communication. You will receive input text and your task is to rephrase it into a short, natural-sounding message written in Gen Z tone.

\textbf{Rules to follow:}
\begin{itemize}
    \item Keep it short and tweet-like.
    \item Use Gen Z slang, abbreviations, exaggerations, or dramatic flair when natural.
    \item Don’t include hashtags, links, or attribution.
    \item Do \textbf{not} include emojis or punctuation.
    \item The top priority is to make the rewritten message sound like something a real Gen Z person might post.
    \item If the text cannot be rewritten (e.g., inappropriate or nonsensical), return ``ERROR\_404''.
\end{itemize}

\textbf{Additional instruction:}  
\texttt{Just output the final text, and nothing else. Do not give pointers or explanations.}
\end{tcolorbox}

\begin{table*}[t]
\centering
\scriptsize 
\renewcommand{\arraystretch}{1.5}
\renewcommand\arraystretch{0.70}
\setlength{\tabcolsep}{5pt}

\begin{tabular}{l l *{3}{c} *{3}{c} *{3}{c} *{3}{c}}
\toprule
\multirow{2}{*}{\makecell[l]{Dimension}} &
\multirow{2}{*}{\makecell[l]{Subgroup}} &
\multicolumn{3}{c}{\textbf{Desklib}} &
\multicolumn{3}{c}{\textbf{E5}} &
\multicolumn{3}{c}{\textbf{Radar}} &
\multicolumn{3}{c}{\textbf{ZipPy}} \\
\cmidrule(lr){3-5}\cmidrule(lr){6-8}\cmidrule(lr){9-11}\cmidrule(lr){12-14}
 & & P & R & F1 & P & R & F1 & P & R & F1 & P & R & F1 \\
\midrule

\multirow{2}{*}{Gender}
  & F & 0.86 & 0.99 & 0.92 & 0.55 & 0.99 & 0.71 & 0.61 & 0.55 & 0.58 & 0.36 & 0.47 & 0.41 \\
  & M & 0.87 & 0.99 & 0.93 & 0.57 & 0.99 & 0.73 & 0.62 & 0.59 & 0.60 & 0.36 & 0.47 & 0.41 \\
\cmidrule(lr){1-14}

\multirow{6}{*}{Race}
  & American Indian  & 0.74 & 0.98 & 0.84 & 0.54 & 0.98 & 0.70 & 0.55 & 0.48 & 0.51 & 0.32 & 0.41 & 0.36 \\
  & API               & 0.84 & 0.99 & 0.91 & 0.55 & 0.99 & 0.71 & 0.56 & 0.47 & 0.51 & 0.38 & 0.51 & 0.44 \\
  & African American  & 0.89 & 0.99 & 0.93 & 0.59 & 0.99 & 0.74 & 0.64 & 0.61 & 0.62 & 0.36 & 0.45 & 0.40 \\
  & Hispanic          & 0.87 & 0.99 & 0.92 & 0.56 & 0.99 & 0.72 & 0.62 & 0.58 & 0.60 & 0.37 & 0.47 & 0.41 \\
  & White             & 0.87 & 0.98 & 0.93 & 0.56 & 0.99 & 0.71 & 0.61 & 0.55 & 0.58 & 0.36 & 0.46 & 0.40 \\
  & Two or more       & 0.86 & 0.99 & 0.92 & 0.55 & 0.99 & 0.71 & 0.64 & 0.53 & 0.58 & 0.33 & 0.40 & 0.36 \\
\cmidrule(lr){1-14}

\multirow{2}{*}{Economic Disadvantage}
  & Y & 0.86 & 0.98 & 0.92 & 0.58 & 0.99 & 0.73 & 0.62 & 0.57 & 0.59 & 0.35 & 0.43 & 0.39 \\
  & N & 0.85 & 0.98 & 0.91 & 0.55 & 0.99 & 0.71 & 0.60 & 0.54 & 0.57 & 0.34 & 0.44 & 0.39 \\
\cmidrule(lr){1-14}

\multirow{2}{*}{Disability Status}
  & Y & 0.84 & 0.99 & 0.91 & 0.61 & 0.99 & 0.76 & 0.63 & 0.59 & 0.61 & 0.34 & 0.39 & 0.36 \\
  & N & 0.86 & 0.99 & 0.92 & 0.55 & 0.99 & 0.71 & 0.63 & 0.57 & 0.60 & 0.37 & 0.50 & 0.43 \\
\cmidrule(lr){1-14}

\multirow{2}{*}{ELL Status}
  & Y & 0.82 & 0.99 & 0.90 & 0.59 & 0.99 & 0.74 & 0.62 & 0.61 & 0.61 & 0.36 & 0.43 & 0.39 \\
  & N & 0.87 & 0.99 & 0.93 & 0.55 & 0.99 & 0.71 & 0.62 & 0.59 & 0.60 & 0.35 & 0.45 & 0.39 \\
\cmidrule(lr){1-14}

\multirow{5}{*}{Grade Level}
  & 8  & 0.90 & 0.98 & 0.94 & 0.57 & 0.99 & 0.72 & 0.64 & 0.78 & 0.71 & 0.41 & 0.68 & 0.51 \\
  & 9  & 0.99 & 0.99 & 0.99 & 0.58 & 0.99 & 0.73 & 0.74 & 0.77 & 0.75 & 0.42 & 0.73 & 0.54 \\
  & 10 & 0.91 & 0.99 & 0.95 & 0.57 & 0.99 & 0.72 & 0.74 & 0.74 & 0.74 & 0.39 & 0.63 & 0.48 \\
  & 11 & 0.93 & 0.99 & 0.96 & 0.55 & 0.99 & 0.71 & 0.69 & 0.74 & 0.72 & 0.42 & 0.69 & 0.52 \\
  & 12 & 0.93 & 0.99 & 0.96 & 0.57 & 0.99 & 0.73 & 0.74 & 0.76 & 0.75 & 0.42 & 0.72 & 0.53 \\
\cmidrule(lr){1-14}

\multirow{4}{*}{Age Level}
  & Teens & 0.86 & 0.49 & 0.62 & 0.57 & 0.60 & 0.59 & 0.19 & 0.17 & 0.18 & 0.21 & 0.01 & 0.02 \\
  & 20s   & 0.82 & 0.53 & 0.64 & 0.56 & 0.77 & 0.65 & 0.24 & 0.25 & 0.24 & 0.36 & 0.01 & 0.03 \\
  & 30s   & 0.81 & 0.60 & 0.69 & 0.55 & 0.83 & 0.66 & 0.30 & 0.32 & 0.31 & 0.42 & 0.03 & 0.05 \\
  & 40s   & 0.76 & 0.62 & 0.68 & 0.54 & 0.86 & 0.67 & 0.35 & 0.40 & 0.38 & 0.57 & 0.05 & 0.09 \\
\cmidrule(lr){1-14}

\multirow{3}{*}{Dialect Bias}
  & Singlish & 0.48 & 0.67 & 0.56 & 0.12 & 0.09 & 0.10 & 0.50 & 0.99 & 0.67 & 0.12 & 0.00 & 0.01 \\
  & AAVE     & 0.45 & 0.66 & 0.54 & 0.43 & 0.02 & 0.04 & 0.50 & 0.99 & 0.67 & 0.19 & 0.00 & 0.01 \\
  & SAE      & 0.57 & 0.88 & 0.69 & 0.01 & 0.00 & 0.00 & 0.50 & 0.99 & 0.67 & 0.05 & 0.00 & 0.00 \\
\cmidrule(lr){1-14}

\multirow{2}{*}{Formality Bias}
  & GenZ     & 0.31 & 0.40 & 0.35 & 0.00 & 0.00 & 0.00 & 0.50 & 0.99 & 0.67 & 0.00 & 0.00 & 0.00 \\
  & Standard & 0.50 & 0.98 & 0.66 & 0.50 & 0.99 & 0.67 & 0.50 & 0.99 & 0.67 & 0.86 & 0.17 & 0.29 \\
\cmidrule(lr){1-14}

\multirow{10}{*}{Topic Level}
  & Arts                & 0.85 & 0.60 & 0.70 & 0.56 & 0.70 & 0.62 & 0.33 & 0.37 & 0.35 & 0.34 & 0.02 & 0.03 \\
  & Communication/Media & 0.82 & 0.69 & 0.75 & 0.55 & 0.79 & 0.65 & 0.35 & 0.41 & 0.38 & 0.51 & 0.04 & 0.07 \\
  & Education           & 0.85 & 0.68 & 0.76 & 0.57 & 0.80 & 0.66 & 0.37 & 0.42 & 0.39 & 0.52 & 0.05 & 0.09 \\
  & Engineering         & 0.85 & 0.66 & 0.74 & 0.57 & 0.75 & 0.65 & 0.35 & 0.39 & 0.37 & 0.62 & 0.05 & 0.09 \\
  & Internet            & 0.81 & 0.61 & 0.69 & 0.54 & 0.79 & 0.64 & 0.35 & 0.41 & 0.38 & 0.57 & 0.04 & 0.08 \\
  & Law                 & 0.87 & 0.76 & 0.81 & 0.55 & 0.84 & 0.67 & 0.35 & 0.40 & 0.37 & 0.74 & 0.07 & 0.13 \\
  & Non-profit          & 0.86 & 0.70 & 0.77 & 0.60 & 0.83 & 0.69 & 0.39 & 0.51 & 0.44 & 0.60 & 0.05 & 0.09 \\
  & Student             & 0.85 & 0.57 & 0.68 & 0.57 & 0.65 & 0.61 & 0.28 & 0.28 & 0.28 & 0.31 & 0.02 & 0.03 \\
  & Technology          & 0.82 & 0.65 & 0.73 & 0.56 & 0.78 & 0.65 & 0.33 & 0.38 & 0.36 & 0.51 & 0.03 & 0.06 \\
  & Unknown             & 0.83 & 0.63 & 0.72 & 0.57 & 0.79 & 0.67 & 0.34 & 0.39 & 0.37 & 0.30 & 0.02 & 0.03 \\
\cmidrule(lr){1-14}

\multirow{3}{*}{Political Ideology}
  & Left leaning  & 0.94 & 0.98 & 0.96 & 0.51 & 0.98 & 0.67 & 0.95 & 0.05 & 0.09 & 0.03 & 0.01 & 0.01 \\
  & Neutral       & 0.90 & 0.99 & 0.94 & 0.50 & 0.98 & 0.66 & 0.96 & 0.05 & 0.09 & 0.05 & 0.01 & 0.02 \\
  & Right leaning & 0.95 & 0.99 & 0.97 & 0.51 & 0.98 & 0.67 & 0.93 & 0.04 & 0.08 & 0.04 & 0.01 & 0.01 \\

\bottomrule
\end{tabular}
\caption{Performance by subgroup across all bias dimensions on AI-generated texts}
\label{tab:all-bias}
\end{table*}

\end{document}